\documentclass[11pt]{article}
\usepackage[final]{acl}

\usepackage{times}
\usepackage{latexsym}
\usepackage[T1]{fontenc} 
\usepackage{microtype}
\usepackage{inconsolata}

\usepackage{fontspec}

\newfontfamily\devanagarifont[
  Path = ./,
  UprightFont = NotoSerifDevanagari-Regular.ttf,
  BoldFont = NotoSerifDevanagari-Bold.ttf,
  Script = Devanagari
]{Noto Serif Devanagari}

\usepackage{graphicx}
\usepackage{amsmath}
\usepackage{amssymb}
\usepackage{amsfonts}
\usepackage{bm}
\usepackage{svg}
\usepackage{float}

\usepackage{booktabs}
\usepackage{multirow}
\usepackage{array}
\usepackage{adjustbox}
\usepackage[table]{xcolor}
\usepackage[most]{tcolorbox}
\usepackage{caption}

\title{Assessing and Improving Punctuation Robustness in English-Marathi Machine Translation}

\author{
  \textbf{Kaustubh Shivshankar Shejole}, 
  \textbf{Sourabh Deoghare} \and 
  \textbf{Pushpak Bhattacharyya} \\
  Computation for Indian Language Technology (CFILT) \\
  Department of Computer Science and Engineering \\
  Indian Institute of Technology Bombay, Mumbai, India \\
  \texttt{\{kaustubhshejole, sourabhdeoghare, pb\}@cse.iitb.ac.in}
}

\begin{document}
\maketitle


\begin{abstract}
Neural Machine Translation (NMT) systems rely heavily on explicit punctuation cues to resolve semantic ambiguities in a source sentence. Inputting user-generated sentences, which are likely to contain missing or incorrect punctuation, results in fluent but semantically disastrous translations. This work attempts to highlight and address the problem of punctuation robustness of NMT systems through an English-to-Marathi translation. First, we introduce \textbf{\textit{Viram}}, a human-curated diagnostic benchmark of 54 punctuation-ambiguous English-Marathi sentence pairs to stress-test existing NMT systems. Second, we evaluate two simple remediation strategies: cascade-based \textit{restore-then-translate} and \textit{direct fine-tuning}. Our experimental results and analysis demonstrate that both strategies yield substantial NMT performance improvements. Furthermore, we find that current Large Language Models (LLMs) exhibit relatively poorer robustness in translating such sentences than these task-specific strategies, thus necessitating further research in this area. The code and dataset are available at \url{https://github.com/KaustubhShejole/Viram_Marathi}.\footnote{Paper accepted at The Ninth Workshop on Technologies for Machine Translation of Low Resource Languages (LoResMT 2026) @ EACL 2026.}
\end{abstract}

\section{Introduction}
Punctuation is an essential component of written language, playing a critical role in resolving both structural and semantic ambiguity. By signaling how textual elements should be grouped and interpreted, punctuation enables readers to accurately infer the intended meaning of a sentence. Broadly, punctuation serves two complementary functions. First, it marks boundaries between segments of a larger statement and encodes grammatical relationships among those segments. Second, it provides rhetorical cues by indicating emphasis, tone, or nuance associated with particular words or phrases \cite{kirkman2006punctuation}.

The importance of punctuation can be illustrated through classic examples. For instance, the omission of a comma in the phrase \textit{“Let’s eat, Grandma.”} transforms an innocent dinner invitation into a cannibalistic implication. Such cases demonstrate how ambiguity naturally arises when punctuation is absent or misused. Similarly, in the sentence “This is known as ‘exact’ recovery.”, quotation marks signal specific emphasis on the term exact, guiding the reader’s interpretation. In general, punctuation errors that affect grammatical structure are more consequential than those that affect rhetorical emphasis as the former can fundamentally alter semantic interpretation \cite{kirkman2006punctuation, carey1980punctuation}.

\begin{figure*}
  \centering
  \includegraphics[width=0.99\linewidth]{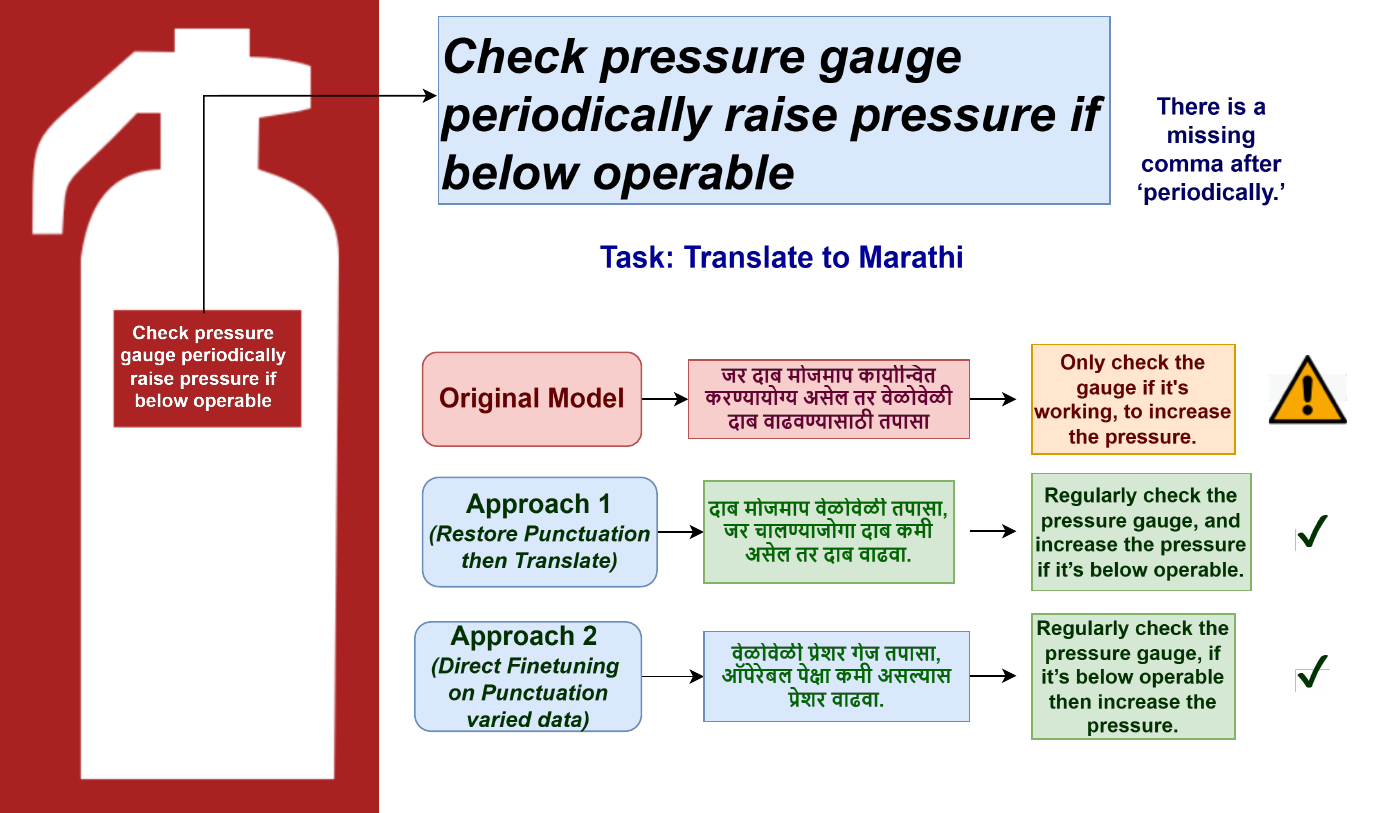}
  \caption{A missing comma can lead to a disaster in English–Marathi machine translation.}
  \label{fig:motivation}
\end{figure*}

The advent of Transformer \cite{vaswani2017attention} has led to rapid improvements in NMT quality over the last few years. Consequently, the applicability of encoder-decoder and Large Language Model (LLM)-based systems has expanded significantly, now encompassing diverse domains and low-resource languages \cite{kocmi-etal-2025-findings, pakray-etal-2025-findings}. In this paper, we focus on Marathi\footnote{\url{https://en.wikipedia.org/wiki/Marathi_language}}, an Indo-Aryan language primarily spoken by over 80 million people in the complex linguistic landscape of India, yet considered a low- to mid-resource language \cite{dabre-etal-2024-machine, lahoti2022survey, gaikwad2021cross}.

Figure \ref{fig:motivation} illustrates an example in which a missing comma in an instruction written on a fire extinguisher could lead to a disaster, highlighting the punctuation sensitivity of NMT systems. Hence, we consider it important to analyze the punctuation sensitivity of current models and to develop techniques to improve their robustness to punctuation, along with an examination of the associated trade-offs. In addition, we emphasize the need to create resources for evaluating punctuation robustness and to explore strategies for improving translation reliability under punctuation variability.

We first describe the data collection process for assessing the punctuation sensitivity of current Indic models, which was carried out via two native speakers of Marathi and a book by \citet{kirkman2006punctuation}, leading to \textbf{\textit{Viram\footnote{\textit{Viram chinhe} ({\devanagarifont विराम चिन्हे}) is a Marathi word for punctuation, i.e., signs for marking boundaries by stopping.}}}, the first English (Written)–English (Meant)–Marathi (Meant) benchmark, where `Meant' refers to the disambiguated semantic representation.
We then apply two approaches for improving the punctuation
robustness of current models, and carry out quantitative and qualitative comparison. We also attempt to evaluate the translation quality via prompting of LLMs. All LLMs we considered exhibit lower performance, indicating the need of punctuation-robust approaches to be developed further. Finally, we analyze the performance of our models on standard benchmarks and observe that on the cost of punctuation robustness
we might lose slightly on evaluation metrics.
Our contributions are as follows:

\begin{enumerate}
\item The first study of punctuation robustness in
English-Marathi machine translation.

\item A novel diagnostic benchmark called \textbf{\textit{Viram}} for English–Marathi punctuation sensitivity analysis. It consists of 54 manually curated, punctuation-ambiguous instances of the form English (Written) – English (Meant) – Marathi (Meant).

\item An analysis of improving punctuation robustness using two complementary approaches: (i) \textit{punctuation restoration in English} then \textit{translate to Marathi}, and (ii) \textit{Direct translation to Marathi}. This dual formulation enables systematic comparison of restoration paradigms. This analysis will help in proliferating further approaches for improving punctuation robustness.

\item A detailed qualitative analysis of model outputs, highlighting strengths, limitations, and error patterns, and identifying directions for future research on punctuation robustness in machine translation.
\end{enumerate}

\begin{table*}[t]
\centering
\resizebox{0.85\textwidth}{!}{
\begin{tabular}{p{0.28\linewidth} p{0.28\linewidth} p{0.28\linewidth} p{0.15\linewidth}}
\toprule
\textbf{English (Written)} & \textbf{English (Meant)} & \textbf{Marathi (Meant)} & \textbf{Punctuation} \\
\midrule
As the machine develops the forms we use to record data from past projects will be amended.
&
As the machine develops\textbf{,} the forms we use to record data from past projects will be amended.
&
{\devanagarifont जसजशी यंत्रणा विकसित होईल, तसतसे मागील प्रकल्पांतील डेटा रेकॉर्ड करण्यासाठी आम्ही वापरत असलेले फॉर्म्स सुधारित केले जातील.}
&
Comma
\\
What we see, we believe what we hear, we register
&
What we see, we believe\textbf{;} what we hear, we register
&
{\devanagarifont जे पाहतो, त्यावर विश्वास ठेवतो; जे ऐकतो, त्याची नोंद घेतो.}
&
Semi Colon
\\
\bottomrule
\end{tabular}
}
\caption{Examples of punctuation ambiguity with English sentences and their Marathi translations in the \textit{Viram} Benchmark}
\label{tab:punctuation_examples}
\end{table*}

\section{Related Work}

Punctuation has long been studied in linguistics for its role in disambiguation, grammatical structure, and rhetorical emphasis \cite{kirkman2006punctuation, carey1980punctuation, lukeman2011art, trask2019penguin}. These works establish how punctuation errors can introduce semantic ambiguity, motivating its importance in downstream language technologies.

In Natural Language Processing (NLP), punctuation restoration has been explored primarily as a preprocessing task for text and speech. Early neural approaches modeled the problem using recurrent architectures, including LSTM-based models \cite{tilk2015lstm} and bidirectional RNNs with attention \cite{tilk2016bidirectional}, particularly for spoken language transcripts. Subsequent work extended punctuation restoration to multilingual and transformer-based settings, including large pretrained models for automatic punctuation and capitalization \cite{nagy2021automatic, puaics2022capitalization}. More recently, systems such as Punktuator \cite{chordia2021punktuator} and Cadence \cite{pulipaka2025mark} have demonstrated robust multilingual and cross-domain punctuation restoration for both text and speech.

Within machine translation, prior studies have acknowledged the role of punctuation in preserving meaning across languages. For example, \citet{mogahed2012punctuation} examined punctuation effects in English–Arabic MT, highlighting its impact on translation quality. However, explicit modeling of punctuation robustness in MT pipelines remains limited.

Recent research on Indic languages has focused on improving translation quality and evaluation, with models like IndicTrans2 \cite{gala2023indictrans2} supporting translation across all 22 scheduled Indian languages, alongside work on MT metric meta-evaluation \cite{dixit2023indicmt} and zero-shot evaluation in low-resource settings \cite{singh2024good}. However, English-to-Marathi translation remains highly sensitive to punctuation cues: standard models such as IndicTrans2 often misinterpret syntactic and semantic relations when punctuation is altered or removed. This highlights a critical gap in current MT systems for Marathi. To address it, we develop punctuation-robust MT models tailored for English–Marathi translation, aiming to improve reliability under punctuation variability.

In contrast to prior work, our study lies at the intersection of punctuation restoration and English–Marathi machine translation. We explicitly examine punctuation sensitivity in MT models and analyze the improvement using punctuation-robust modeling approaches, addressing a gap in both Indic MT and punctuation restoration literature.

\section{Creating the \textit{Viram} Benchmark}
\citet{kirkman2006punctuation} analyze punctuation in the English language, examining how ambiguity can arise from the omission of punctuation marks. For instance, in the sentence, ``As the machine develops the forms we use to record data from past projects will be amended,'' readers must insert a comma after \emph{develops} to derive the intended meaning. This example illustrates the human ability to extract meaning from syntactically ambiguous sentences. 
Given that \citet{kirkman2006punctuation} is a well-established resource, we manually curated English sentences from this work and, with the assistance of two native Marathi speakers, translated the English (Meant) sentences into Marathi. The resulting diagnostic benchmark comprises 54 punctuation-ambiguous instances, structured as English (Written) – English (Meant) – Marathi (Meant). 
While the benchmark size is relatively modest, it is commensurate with the significant challenges inherent in data acquisition and curation within this specific domain. Despite this, the rigor applied to its curation ensures that it serves as a high-quality, representative sample for diagnostic evaluation. Table~\ref{tab:punctuation_examples} presents selected examples from \textit{Viram}, illustrating the nature of punctuation ambiguities and their corresponding translations.
Details regarding the annotation process are provided in Appendix~\ref{app:sec_viram_more_details}.

\section{Methodology}
\label{sec:approaches}
We explore two primary paradigms for achieving punctuation robustness in English-to-Marathi translation.

\subsection{Approach 1: \textit{Restore Punctuation} \textit{then} \textit{Translate}}
In this decouple-and-conquer approach, punctuation is first restored in the English source text before translation, reducing the task to punctuation restoration. 
We adopt two modeling paradigms for punctuation restoration. In the \textbf{token classification} approach, \texttt{bert-large-uncased} \cite{devlin2019bert} and \texttt{microsoft-mpnet-base} \cite{song2020mpnet} are used to treat punctuation prediction as a sequence labeling task. In the \textbf{text-to-text generation} approach, we fine-tune \texttt{google-t5-base} \cite{2020t5} to generate punctuated text from unpunctuated input and also evaluate AI4Bharat’s \texttt{Cadence} model ~\cite{pulipaka2025mark} without fine-tuning it.

\subsection{Approach 2: \textit{Direct Translation}}

This approach aims to improve MT robustness to noisy input. We fine-tune the \texttt{IndicTrans2} model\footnote{\url{https://huggingface.co/ai4bharat/indictrans2-en-indic-dist-200M}}
 on four variants of our internal dataset\footnote{It is an in-house corpus created by professional human translators as part of another project. The internal dataset details are provided at \url{https://github.com/KaustubhShejole/Viram_Marathi}}.
We construct four variants using the original data with punctuation (\textbf{With Punct}) as a baseline, removing all source punctuation (\textbf{Without Punct}), combining both original and punctuation-removed data (\textbf{Combined 2x}), and alternately retaining or removing punctuation on a per-sentence basis (\textbf{Combined x}).
Please note that `x' refers to the size of the internal fine-tuning dataset.
Details for data handling are provided  for both the approaches in Appendix~\ref{app:approach_1_data} and ~\ref{app:approach_2_data} respectively.
Details regarding fine-tuning and hyperparameter selection are provided in Appendix \ref{app:fine-tune-hyper}.

\begin{table*}[t]
\centering

\renewcommand{\arraystretch}{1.41}

\setlength{\tabcolsep}{6pt}

\begin{adjustbox}{max width=\textwidth}
\large
\begin{tabular}{llccccccc}
\toprule
\textbf{Type} & \textbf{Model Name} & \textbf{BLEU} & \textbf{BLEURT-20} & \textbf{COMET} & \textbf{chrF++} & \textbf{chrF2++} & \textbf{LabSE} & \textbf{MuRIL} \\
\midrule

Baseline &
IndicTrans2 en indic 200M (Original Model) &
21.72 & 0.7916 & 0.7391 & 59.45 & 55.38 & 0.9126 & 0.7619 \\
\midrule

Upper Performance Boundary &
IndicTrans2 en indic 200M + Input as `sentence meant' &
\textbf{26.20} & \textbf{0.8082} & \textbf{0.7606} & \textbf{61.15} & \textbf{57.41} & \textbf{0.9313} & \textbf{0.7915} \\
\midrule

\multirow{4}{*}{Approach 1} &
Fine-tuned bert-large-uncased + Original Model &
23.84 & 0.7955 & 0.7595 & 60.02 & 56.11 & 0.9199 & 0.7806 \\

& Fine-tuned microsoft-mpnet + Original Model &
\textbf{25.12} & \textbf{0.7996} & \textbf{0.7597} & 60.56 & 56.79 & 0.9210 & 0.7813 \\

& Fine-tuned t5-base + Original Model &
24.74 & 0.7977 & 0.7586 & \textbf{60.68} & \textbf{56.92} & \textbf{0.9230} & \textbf{0.7838} \\

& AI4Bharat’s cadence + Original Model &
23.44 & 0.7980 & 0.7516 & 60.49 & 56.69 & 0.9210 & 0.7809 \\
\midrule

\multirow{4}{*}{Approach 2} &
Finetuned (w/ punct) (x) &
21.21 & \textbf{0.7830} & 0.7426 & 58.90 & 54.72 & \textbf{0.9145} & 0.7685 \\

& Finetuned (w/o punct) (x) &
24.66 & 0.7774 & 0.7417 & 60.30 & 56.56 & 0.9122 & \textbf{0.7830} \\

& Finetuned (with and w/o punct) (2x) &
24.27 & 0.7785 & \textbf{0.7443} & \textbf{60.61} & \textbf{56.83} & 0.9120 & 0.7794 \\

& Finetuned (alternate with and w/o punct) (x) &
24.28 & 0.7745 & 0.7433 & 60.21 & 56.52 & 0.9047 & 0.7761 \\
\midrule

\multirow{2}{*}{LLM} &
GPT-5-mini (Zero-Shot + Direct Translation) &
18.69 & 0.7786 & 0.7420 & 52.50 & 48.82 & 0.9096 & 0.7394 \\

& DeepSeek-V3.2 (Zero-Shot + Direct Translation) &
\textbf{23.41} & \textbf{0.7858} & \textbf{0.7590} & \textbf{58.48} & \textbf{54.82} & \textbf{0.9197} & \textbf{0.7765} \\

\bottomrule
\end{tabular}
\end{adjustbox}

\caption{Quantitative Analysis on the \textit{Viram} Benchmark}
\label{tab:quantitative_analysis}
\end{table*}

\begin{table*}[t]
\centering

\renewcommand{\arraystretch}{1.35}

\setlength{\tabcolsep}{6pt}

\small
\begin{adjustbox}{max width=0.99\textwidth}
\begin{tabular}{llccccccc}
\toprule
\textbf{Benchmark} & \textbf{Model Name} & \textbf{BLEU} & \textbf{BLEURT-20} & \textbf{COMET} & \textbf{chrF++} & \textbf{chrF2++} & \textbf{LabSE} & \textbf{MuRIL} \\
\midrule
IN22 (CONV) & IndicTrans2 en indic 200 M (Original) & 18.95 & 0.8209 & 0.8117 & 51.05 & 47.74 & 0.9051 & 0.7451 \\
            & Fine-tuned t5-base + Original Model & 17.82 & 0.8121 & 0.8077 & 50.24 & 47.17 & 0.8982 & 0.7404 \\
            & Finetuned (w/ punct) (x) & 16.08 & 0.7963 & 0.7920 & 49.77 & 46.65 & 0.8869 & 0.7300 \\
            & Finetuned (w/o punct) (x) & 16.67 & 0.8034 & 0.7995 & 49.68 & 46.72 & 0.8927 & 0.7354 \\
            & Finetuned (with and w/o punct) (2x) & 17.93 & 0.8106 & 0.8065 & 50.45 & 47.33 & 0.8962 & 0.7408 \\
            & Finetuned (alternate with and w/o punct) (x) & 17.87 & 0.8122 & 0.8082 & 50.34 & 47.26 & 0.8983 & 0.7411 \\
\midrule
IN22 (GEN)  & IndicTrans2 en indic 200 M (Original) & 21.01 & 0.7920 & 0.7539 & 54.22 & 49.99 & 0.9153 & 0.7389 \\
            & Fine-tuned t5-base + Original Model & 16.86 & 0.7819 & 0.7439 & 53.19 & 48.84 & 0.9129 & 0.7309 \\
            & Finetuned (w/ punct) (x) & 16.98 & 0.7627 & 0.7405 & 53.23 & 48.86 & 0.9098 & 0.7290 \\
            & Finetuned (w/o punct) (x) & 17.07 & 0.7745 & 0.7422 & 53.09 & 48.77 & 0.9116 & 0.7307 \\
            & Finetuned (with and w/o punct) (2x) & 17.11 & 0.7809 & 0.7440 & 53.50 & 49.12 & 0.9131 & 0.7315 \\
            & Finetuned (alternate with and w/o punct) (x) & 17.06 & 0.7822 & 0.7450 & 53.37 & 48.99 & 0.9132 & 0.7312 \\
\midrule
FLORES-22   & IndicTrans2 en indic 200 M (Original) & 18.69 & 0.7894 & 0.7616 & 54.48 & 50.19 & 0.9226 & 0.7425 \\
            & Fine-tuned t5-base + Original Model & 19.34 & 0.7818 & 0.7531 & 55.02 & 50.82 & 0.9213 & 0.7413 \\
            & Finetuned (w/ punct) (x) & 19.20 & 0.7786 & 0.7513 & 54.78 & 50.57 & 0.9198 & 0.7400 \\
            & Finetuned (w/o punct) (x) & 19.27 & 0.7795 & 0.7528 & 54.73 & 50.57 & 0.9201 & 0.7413 \\
            & Finetuned (with and w/o punct) (2x) & 19.37 & 0.7810 & 0.7521 & 55.11 & 50.90 & 0.9204 & 0.7418 \\
            & Finetuned (alternate with and w/o punct) (x) & 19.46 & 0.7825 & 0.7533 & 55.11 & 50.90 & 0.9218 & 0.7421 \\
\bottomrule
\end{tabular}
\end{adjustbox}
\caption{Performance Comparison across Benchmark Datasets}
\label{tab:performance_results}
\end{table*}

\begin{table*}
\centering

\renewcommand{\arraystretch}{1.5}

\setlength{\tabcolsep}{6pt}

\begin{adjustbox}{max width=0.99\textwidth}
\begin{tabular}{lllccccccc}
\toprule
\textbf{Prompting Strategy} & \textbf{Approach} & \textbf{Model Name} & \textbf{BLEU} & \textbf{BLEURT-20} & \textbf{COMET} & \textbf{chrF++} & \textbf{chrF2++} & \textbf{LabSE} & \textbf{MuRIL} \\ \midrule

\multirow{6}{*}{Zero-Shot} & \multirow{3}{*}{Restore then Translate} 
 & Llama 3.1 8b & 5.16 & 0.6548 & 0.6026 & 37.13 & 33.18 & 0.8030 & 0.6093 \\ 
 & & Gemma 2 9b & 11.69 & 0.7260 & 0.6783 & 45.89 & 41.98 & 0.8783 & 0.6952 \\ 
 & & Sarvam 2b v0.5 & 9.97 & 0.6961 & 0.6736 & 42.93 & 38.86 & 0.8228 & 0.6468 \\ \cmidrule{2-10}
 & \multirow{3}{*}{Direct Translation} 
 & Llama 3.1 8b & 6.37 & 0.6660 & 0.6192 & 39.89 & 35.76 & 0.8230 & 0.6314 \\ 
 & & Gemma 2 9b & 8.94 & 0.7248 & 0.6880 & 45.40 & 41.33 & 0.8559 & 0.6712 \\ 
 & & Sarvam 2b v0.5 & 5.67 & 0.6300 & 0.6392 & 37.47 & 33.67 & 0.8102 & 0.6149 \\ \midrule

\multirow{6}{*}{3-Shot Prompting} & \multirow{3}{*}{Restore then Translate} 
 & Llama 3.1 8b & 4.56 & 0.6641 & 0.6113 & 37.80 & 33.68 & 0.8194 & 0.6120 \\ 
 & & Gemma 2 9b & 14.84 & 0.7407 & 0.6889 & 49.27 & 45.50 & 0.8926 & 0.7056 \\ 
 & & Sarvam 2b v0.5 & 8.55 & 0.7352 & 0.6982 & 44.48 & 40.02 & 0.8509 & 0.6710 \\ \cmidrule{2-10}
 & \multirow{3}{*}{Direct Translation} 
 & Llama 3.1 8b & 7.98 & 0.6776 & 0.6272 & 43.21 & 38.91 & 0.8216 & 0.6192 \\ 
 & & Gemma 2 9b & 10.99 & 0.7388 & 0.6938 & 48.98 & 44.88 & 0.8883 & 0.7013 \\ 
 & & Sarvam 2b v0.5 & 11.01 & 0.7571 & 0.7224 & 49.27 & 44.74 & 0.8803 & 0.7019 \\ \midrule

\multirow{3}{*}{\shortstack[l]{Direct Translation using\\Sentence Meant (Original)}} 
 & \multirow{3}{*}{Direct Translation} 
 & Llama 3.1 8b & 7.87 & 0.6785 & 0.6330 & 39.70 & 35.84 & 0.8404 & 0.6535 \\ 
 & & Gemma 2 9b & 13.75 & 0.7256 & 0.6941 & 45.85 & 42.47 & 0.8829 & 0.6896 \\ 
 & & Sarvam 2b v0.5 & 10.36 & 0.6866 & 0.6725 & 41.30 & 37.63 & 0.8422 & 0.6550 \\ \bottomrule

\end{tabular}
\end{adjustbox}
\caption{Quantitative Analysis of LLMs via various prompting strategies on the \textit{Viram} Benchmark}
\label{tab:llm_quant_analysis}
\end{table*}

\section{Prompting LLMs}

We attempt to evaluate the translation quality of punctuation-ambiguous sentences from English to Marathi using zero-shot and few-shot prompting across three LLMs. The models considered are Sarvam-2b-v0.5\footnote{\url{https://huggingface.co/sarvamai/sarvam-2b-v0.5}}, a 2-billion-parameter model; Gemma-2-9b\footnote{\url{https://huggingface.co/google/gemma-2-9b}} \cite{gemma_2024}, a 9-billion-parameter model; and LLaMA-3.1-8b\footnote{\url{https://huggingface.co/meta-llama/Llama-3.1-8B}} \cite{grattafiori2024llama}, an 8-billion-parameter model. All three models have been exposed to Indian languages during pre-training. Notably, Sarvam-2b-v0.5 has been trained exclusively on English and Indian languages, including Marathi, using a corpus of approximately one trillion tokens per language.
We adopt the same methodology described in Section~\ref{sec:approaches} for each prompting strategy:

\begin{enumerate}
    \item \textbf{Zero-shot prompting}
    \begin{enumerate}
        \item Restore punctuation, then translate (see Appendix~\ref{app:prompt_zero_chain} for details about the prompt).
        \item Direct translation (see Appendix~\ref{app:prompt_zero_zero} for details about the prompt).
    \end{enumerate}
    
    \item \textbf{Three-shot prompting}
    \begin{enumerate}
        \item Restore punctuation, then translate (see Appendix~\ref{app:prompt_chain_chain} for details about the prompt).
        \item Direct translation (see Appendix~\ref{app:prompt_chain_zero} for details about the prompt).
    \end{enumerate}
\end{enumerate}

For outputs using correctly punctuated inputs (original sentence-meant), we employed the direct translation strategy in Appendix~\ref{app:prompt_original}. For three-shot prompting, we selected three examples from the Viram benchmark, each illustrating a distinct punctuation error involving commas, semicolons, and colons. During evaluation, these examples were excluded from the test set to ensure a fair and unbiased assessment.

\section{Results and Analysis}

In this section, we present a comprehensive evaluation of the proposed approaches. We first provide a quantitative comparison of all methods (\S\ref{subsec:quantitative}). We then analyze the impact of the two proposed improvement strategies on the original model’s performance across standard benchmarks (\S\ref{subsec:perf_standard}). Next, we examine the effectiveness of different prompting strategies for large language models (\S\ref{sec:analysis_prompt}). Finally, we complement the quantitative results with a qualitative analysis to better understand the strengths and limitations of the models (\S\ref{sec:qualitative_analysis}).

\subsection{Quantitative Performance}
\label{subsec:quantitative}
Table~\ref{tab:quantitative_analysis} reports quantitative results on the \textit{Viram} benchmark across lexical, semantic, and embedding-based metrics. Details about metrics are provided in Appendix \ref{app:metrics}. The original model with `sent-written' input serves as the baseline, while providing oracle sentence boundaries establishes an upper bound, yielding substantial gains in BLEU, chrF, and embedding similarity. This gap highlights the impact of correct sentence boundary recovery on translation quality.
Pipeline-based punctuation restoration (Approach~1) consistently outperforms the baseline, with \texttt{t5-base} and \texttt{mpnet} restorers approaching the oracle upper bound, indicating that higher-quality punctuation directly improves translation. Direct fine-tuning (Approach~2) on unpunctuated data yields clear gains in BLEU and chrF, while as expected, training only on punctuated data offers limited improvement. Mixed training improves robustness, particularly in COMET and chrF, but still falls short of the strongest pipeline-based results.

Among LLMs, DeepSeek-V3.2 outperforms GPT 5-mini across all metrics, achieving competitive semantic similarity scores in a zero-shot setting. However, both LLMs remain below the strongest pipeline and oracle-segmentation configurations. Overall, accurate sentence boundary recovery is critical for translation quality on \textit{Viram}, with pipeline-based restoration most effective when segmentation quality is high, while fine-tuning improves robustness to punctuation variability.


\subsection{Performance Analysis on Standard Benchmarks}
\label{subsec:perf_standard}
Table~\ref{tab:performance_results} reports automatic evaluation results on IN22 (CONV)\footnote{\url{https://huggingface.co/datasets/ai4bharat/IN22-Conv}}, IN22 (GEN)\footnote{\url{https://huggingface.co/datasets/ai4bharat/IN22-Gen}}, and FLORES-22\footnote{\url{https://indictrans2-public.objectstore.e2enetworks.net/flores-22_dev.zip}}. We compare the original model with pipeline-based punctuation restoration (Approach~1) and direct fine-tuning variants on punctuated, unpunctuated, and mixed data (Approach~2).

On IN22 (CONV), the original model achieves the highest BLEU, while fine-tuned variants show modest drops. Models trained on both punctuated and unpunctuated data outperform single-input variants, with the alternate mixed strategy narrowing the gap with the original on BLEURT-20, COMET, LabSE, and MuRIL.
On IN22 (GEN), the original model again outperforms fine-tuned variants. Pipeline-based punctuation restoration harms BLEU and semantic scores, indicating error propagation. Mixed fine-tuning outperforms single-condition models but remains below the original.
On FLORES-22, all models perform similarly, with some fine-tuned variants slightly exceeding the original in BLEU and chrF without reducing semantic scores.
This suggests that gains in punctuation robustness may come at the cost of slight reductions in certain evaluation metrics.

\begin{table*}[htbp]
\centering
\scriptsize 
\resizebox{0.97\textwidth}{!}{
\begin{tabular}{p{0.15\linewidth}>{\raggedright\arraybackslash}p{0.1\linewidth}p{0.175\linewidth}p{0.3\linewidth}p{0.175\linewidth}>{\raggedright\arraybackslash}p{0.1\linewidth}}
\toprule
\textbf{Sentence (English Input)} & \textbf{Model} & \textbf{Marathi Output} & \textbf{Word-by-Word [English Gloss]} & \textbf{Inferred Meaning} & \textbf{Status} \\ \midrule
1a. Police say suspect confessed lawyer denies claim & Original & {\devanagarifont संशयिताच्या वकिलाने दावा फेटाळल्याचे पोलिसांचे म्हणणे} & {\devanagarifont संशयिताच्या} (Suspect's) {\devanagarifont वकिलाने} (by lawyer) {\devanagarifont दावा} (claim) {\devanagarifont फेटाळल्याचे} (denied) {\devanagarifont पोलिसांचे} (Police's) {\devanagarifont म्हणणे} (saying) & Police say the suspect's lawyer denied the claim. (Missing the confession part of suspect).& Incorrect \\ \cmidrule{2-6}
 & Approach 2 (with and w/o punct) (2x) & {\devanagarifont पोलिसांचे म्हणणे आहे की संशयिताने कबूल केलेला दावा वकील नाकारतो.} & {\devanagarifont पोलिसांचे} (Police's) {\devanagarifont म्हणणे} (say) {\devanagarifont आहे} (is) {\devanagarifont की} (that) {\devanagarifont संशयिताने} (by suspect) {\devanagarifont कबूल} (confessed) {\devanagarifont केलेला} (done) {\devanagarifont दावा} (claim) {\devanagarifont वकील} (lawyer) {\devanagarifont नाकारतो} (denies). & The lawyer denies the claim that the suspect confessed. & Correct\\ \cmidrule{2-6}
 & Approach 1 (t5-base) & {\devanagarifont पोलिसांचे म्हणणे आहे की संशयिताने कबूल केलेला वकील दावा नाकारतो.} & {\devanagarifont पोलिसांचे} (Police's) {\devanagarifont म्हणणे} (say) {\devanagarifont आहे} (is) {\devanagarifont की} (that) {\devanagarifont संशयिताने} (by suspect) {\devanagarifont कबूल} (confessed) {\devanagarifont केलेला} (done) {\devanagarifont वकील} (lawyer) {\devanagarifont दावा} (claim) {\devanagarifont नाकारतो} (denies). & Police say that the claim is denied by the lawyer that the suspect confessed.& Correct\\ \midrule

1b. Police say suspect confessed, lawyer denies claim. & All Models & {\devanagarifont पोलिसांनी सांगितले की संशयिताने कबुली दिली, वकील दावा नाकारतो.} & {\devanagarifont पोलिसांनी} (Police) {\devanagarifont सांगितले} (said) {\devanagarifont की} (that) {\devanagarifont संशयिताने} (suspect) {\devanagarifont कबुली} (confession) {\devanagarifont दिली} (gave), {\devanagarifont वकील} (lawyer) {\devanagarifont दावा} (claim) {\devanagarifont नाकारतो} (denies). & Two separate reports: one confession, one denial. & Correct \\ \midrule

2a. Minister says reform failed opposition celebrates & Original & {\devanagarifont सुधारणांमध्ये अपयशी ठरलेल्या विरोधी पक्षांचा जल्लोषः मंत्री} & {\devanagarifont सुधारणांमध्ये} (In reforms) {\devanagarifont अपयशी} (failed) {\devanagarifont ठरलेल्या} (proven) {\devanagarifont विरोधी} (opposition) {\devanagarifont पक्षांचा} (parties') {\devanagarifont जल्लोष} (celebration): {\devanagarifont मंत्री} (Minister) & The Minister notes the celebration of the opposition failed in reforms. & Incorrect \\ \cmidrule{2-6}
 & Approach 2 (with and w/o punct) (2x) & {\devanagarifont मंत्री म्हणतात, सुधारणा अयशस्वी झाल्याबद्दल विरोधकांनी जल्लोष केला.} & {\devanagarifont मंत्री} (Minister) {\devanagarifont म्हणतात} (says), {\devanagarifont सुधारणा} (reform) {\devanagarifont अयशस्वी} (unsuccessful) {\devanagarifont झाल्याबद्दल} (about becoming) {\devanagarifont विरोधकांनी} (by opposition) {\devanagarifont जल्लोष} (celebration) {\devanagarifont केला} (did). & The Minister says the opposition celebrated the failure of reforms. & Correct \\ \cmidrule{2-6}
 & Approach 1 (t5-base) & {\devanagarifont मंत्री म्हणतात की सुधारणा अयशस्वी झाल्या, विरोधक जल्लोष करतात.} & {\devanagarifont मंत्री} (Minister) {\devanagarifont म्हणतात} (says) {\devanagarifont की} (that) {\devanagarifont सुधारणा} (reforms) {\devanagarifont अयशस्वी} (unsuccessful) {\devanagarifont झाल्या} (became), {\devanagarifont विरोधक} (opposition) {\devanagarifont जल्लोष} (celebration) {\devanagarifont करतात} (do). & Minister says reforms failed and the opposition celebrates. & Correct \\ \midrule

2b. Minister says reform failed, opposition celebrates. & All models & {\devanagarifont मंत्री म्हणतात की सुधारणा अयशस्वी झाल्या, विरोधक जल्लोष करतात.} & (Same as Approach 1 (t5-base) above) & (Same as Approach 1 (t5-base) above) & Correct \\ 

\midrule

3a. Check pressure gauge periodically raise pressure if below operable & Original & {\devanagarifont जर दाब मोजमाप कार्यान्वित करण्यायोग्य असेल तर वेळोवेळी दाब वाढवण्यासाठी तपासा} & {\devanagarifont जर} (If) {\devanagarifont दाब} (pressure) {\devanagarifont मोजमाप} (gauge) {\devanagarifont कार्यान्वित} (operable) {\devanagarifont असेल} (is) {\devanagarifont तर} (then) {\devanagarifont वेळोवेळी} (periodically) {\devanagarifont दाब} (pressure) {\devanagarifont वाढवण्यासाठी} (to increase) {\devanagarifont तपासा} (check). & Only check the gauge if it's working, to increase pressure. & Incorrect \\ \cmidrule{2-6}
 & Approach 2 (with and w/o punct) (2x) & {\devanagarifont प्रेशर गेज वेळोवेळी तपासा, जर ऑपेरेबल पेक्षा कमी असेल तर प्रेशर वाढवा.} & {\devanagarifont प्रेशर गेज} (Pressure gauge) {\devanagarifont वेळोवेळी} (periodically) {\devanagarifont तपासा} (check), {\devanagarifont जर} (if) {\devanagarifont ऑपेरेबल} (operable) {\devanagarifont पेक्षा} (than) {\devanagarifont कमी} (less) {\devanagarifont असेल} (is) {\devanagarifont तर} (then) {\devanagarifont प्रेशर} (pressure) {\devanagarifont वाढवा} (increase). & Regular checks; increase pressure only if it's too low. & Correct \\ \cmidrule{2-6}
 & Approach 1 (t5-base) & {\devanagarifont दाब मोजमाप वेळोवेळी तपासा, जर चालण्याजोगा दाब कमी असेल तर दाब वाढवा.} & {\devanagarifont दाब} (Pressure) {\devanagarifont मोजमाप} (gauge) {\devanagarifont वेळोवेळी} (periodically) {\devanagarifont तपासा} (check), {\devanagarifont जर} (if) {\devanagarifont चालण्याजोगा} (operable) {\devanagarifont दाब} (pressure) {\devanagarifont कमी} (low) {\devanagarifont असेल} (is) {\devanagarifont तर} (then) {\devanagarifont दाब} (pressure) {\devanagarifont वाढवा} (increase). & Check the gauge; if pressure is not operable, increase it. & Correct \\ \midrule

3b. Check pressure gauge periodically, raise pressure if below operable. & Original & {\devanagarifont दाब मोजमाप वेळोवेळी तपासा, जर चालण्याजोगा दाब कमी असेल तर दाब वाढवा.} & (Same as Approach 1 (t5-base) above) & (Same as above) & Correct \\ \cmidrule{2-6}
 & Approach 2 (with and w/o punct) (2x) & {\devanagarifont वेळोवेळी प्रेशर गेज तपासा, ऑपेरेबल पेक्षा कमी असल्यास प्रेशर वाढवा.} & {\devanagarifont वेळोवेळी} (Periodically) {\devanagarifont प्रेशर गेज} (gauge) {\devanagarifont तपासा} (check), {\devanagarifont ऑपेरेबल} (operable) {\devanagarifont पेक्षा} (than) {\devanagarifont कमी} (less) {\devanagarifont असल्यास} (if being) {\devanagarifont प्रेशर} (pressure) {\devanagarifont वाढवा} (increase). & (Same as above) & Correct \\ \cmidrule{2-6}
 & Approach 1 (t5-base) & {\devanagarifont दाब मोजमाप वेळोवेळी तपासा, जर चालण्याजोगा दाब कमी असेल तर दाब वाढवा.} & (Same as Approach 1 (t5-base) above) & (Same as above) & Correct \\

\midrule

4a. What we see we believe what we hear we register & Original & {\devanagarifont आपण जे पाहतो त्यावर आपण विश्वास ठेवतो की आपण जे ऐकतो त्यावर आपण नोंदणी करतो.} & {\devanagarifont आपण} (We) {\devanagarifont जे} (what) {\devanagarifont पाहतो} (see) {\devanagarifont त्यावर} (on that) {\devanagarifont आपण} (we) {\devanagarifont विश्वास} (believe) {\devanagarifont ठेवतो} (keep) {\devanagarifont की} (OR) {\devanagarifont आपण} (we) {\devanagarifont जे} (what) {\devanagarifont ऐकतो} (hear) {\devanagarifont त्यावर} (on that) {\devanagarifont आपण} (we) {\devanagarifont नोंदणी} (register) {\devanagarifont करतो} (do). & A choice: Do we believe what we see OR register what we hear? & Incorrect \\ \cmidrule{2-6}
 & Approach 2 (with and w/o punct) (2x) & {\devanagarifont आपण जे पाहतो त्यावर आपण विश्वास ठेवतो, आपण जे ऐकतो त्यावर आपण नोंदणी करतो.} & {\devanagarifont आपण} (We) {\devanagarifont जे} (what) {\devanagarifont पाहतो} (see) {\devanagarifont त्यावर} (on that) {\devanagarifont आपण} (we) {\devanagarifont विश्वास} (believe) {\devanagarifont ठेवतो} (keep), {\devanagarifont आपण} (we) {\devanagarifont जे} (what) {\devanagarifont ऐकतो} (hear) {\devanagarifont त्यावर} (on that) {\devanagarifont आपण} (we) {\devanagarifont नोंदणी} (register) {\devanagarifont करतो} (do). & We believe what we see, and we register what we hear. & Correct \\ \cmidrule{2-6}
 & Approach 1 (t5-base) & {\devanagarifont आपण जे पाहतो, विश्वास ठेवतो, जे ऐकतो, ते नोंदवतो.} & {\devanagarifont आपण} (We) {\devanagarifont जे} (what) {\devanagarifont पाहतो} (see), {\devanagarifont विश्वास} (believe) {\devanagarifont ठेवतो} (keep), {\devanagarifont जे} (what) {\devanagarifont ऐकतो} (hear), {\devanagarifont ते} (that) {\devanagarifont नोंदवतो} (register). & Seeing leads to belief, hearing leads to registration. & Correct \\ \midrule

4b. What we see we believe; what we hear we register. & All Models & {\devanagarifont आपण जे पाहतो त्यावर आपण विश्वास ठेवतो; जे ऐकतो त्यावर आपण नोंदणी करतो.} & {\devanagarifont आपण} (We) {\devanagarifont जे} (what) {\devanagarifont पाहतो} (see) {\devanagarifont त्यावर} (on that) {\devanagarifont आपण} (we) {\devanagarifont विश्वास} (believe) {\devanagarifont ठेवतो} (keep); {\devanagarifont जे} (what) {\devanagarifont ऐकतो} (hear) {\devanagarifont त्यावर} (on that) {\devanagarifont आपण} (we) {\devanagarifont नोंदणी} (register) {\devanagarifont करतो} (do). & Parallel statements of two human actions. & Correct \\

\bottomrule
\end{tabular}
}
\caption{Qualitative comparison of translation outputs of the original models and fine-tuned models on two approaches.}
\label{tab:qualitative1}
\end{table*}

\subsection{Analysis of Prompting Strategies in LLMs}
\label{sec:analysis_prompt}
The results in Table~\ref{tab:llm_quant_analysis} show that, among zero-shot prompting strategies, Gemma 2 9B consistently outperforms the other evaluated LLMs across most metrics. When considering all prompting strategies, LLaMA 3.1 8B exhibits comparatively lower performance than Gemma 2 and Sarvam 2B, highlighting the impact of model architecture and pretraining scale on multilingual translation quality. In zero-shot settings, Sarvam performs better under Approach 1, whereas the other models achieve higher scores with Approach 2. Under 3-shot prompting, both LLaMA and Sarvam benefit more from Approach 2, while Gemma continues to achieve superior results with Approach 1. 

When these LLM results are compared to the quantitative baselines reported in Table~\ref{tab:quantitative_analysis}, it becomes apparent that sub-10B parameter models generally underperform relative to closed-source models such as DeepSeek-V3.2 and GPT 5-mini, which benefit from more specialized capabilities. For instance, DeepSeek-V3.2 achieves a BLEU score of 23.41 and a BLEURT-20 score of 0.7858, whereas GPT 5-mini in a zero-shot direct translation setting attains BLEU 18.69 and BLEURT-20 0.7786. In contrast, Gemma 2 9b, the best-performing model under zero-shot and three-shot prompting conditions, reaches a BLEU score of 14.84 and BLEURT-20 of 0.7407. These results further suggest that targeted fine-tuning or augmented input strategies continue to offer substantially higher translation quality, particularly on metrics that are sensitive to semantic adequacy, such as COMET and LabSE.

\subsection{Qualitative Analysis}
\label{sec:qualitative_analysis}
Tables~\ref{tab:qualitative1} presents a qualitative comparison of translations produced by the original model, Approach~1 (punctuation restoration using T5-base followed by translation), and Approach~2 (Combined 2x: direct fine-tuning on punctuated and unpunctuated data). The examples evaluate the models’ ability to resolve syntactic ambiguity, clause boundaries, and semantic scope in punctuation-sparse inputs.
The original model consistently struggles with unpunctuated sentences, particularly in headline-style constructions (e.g., 1a, 2a) and instructional text (3a). In news headlines containing multiple reporting verbs, the model frequently misidentifies clause attachment and argument scope, either omitting one of the reported events (1a) or incorrectly attributing actions to the wrong entity (2a). In procedural sentences, it often embeds conditional phrases incorrectly, conflating the condition with the action itself rather than expressing a sequence of operations (3a). Similarly, in parallel or contrastive constructions (4a), the absence of punctuation leads the model to misinterpret coordination as disjunction, resulting in unintended semantic alternation. These errors indicate a strong dependence on explicit punctuation cues for recovering sentence structure.

Approach~1 substantially improves translation quality by restoring punctuation prior to translation. This enables more accurate recovery of clause boundaries, coordination, and reporting structures, yielding correct interpretations in most ambiguous cases (e.g., 1a, 2a, 4a). However, as a pipeline approach, it remains sensitive to errors introduced during punctuation restoration, which can occasionally propagate into the final translation and lead to less natural or slightly misaligned syntactic realizations.
Approach~2 consistently produces the most accurate and stable translations across all examples. The fine-tuned model correctly resolves implicit coordination, reporting structures, and conditional logic even in the absence of punctuation, as demonstrated in both news headlines and procedural instructions. Performance remains robust across punctuated and unpunctuated variants, suggesting that the model learns to infer latent sentence structure directly from contextual and syntactic cues rather than relying on surface punctuation.

For punctuated inputs (b variants), all models produce correct translations, confirming that most observed errors in the original model arise from difficulties in handling missing punctuation rather than lexical or morphological limitations. 
These results demonstrate that fine-tuning may help in combating the punctuation-sensitivity of the original model for English-Marathi machine translation.
While automatic metrics show moderate gains, qualitative evaluation reveals substantial improvements in semantic fidelity that are not captured by standard scores.

\section{Conclusion and Future Directions}
In this study, we focused on assessing and improving punctuation robustness of English–to-Marathi NMT systems. We manually constructed \textbf{\textit{Viram}}, a diagnostic benchmark that contains punctuation-ambiguous instances. We evaluated two primary approaches: a pipeline-based \textit{restore-then-translated} and \textit{direct fine-tuning} on punctuated and unpunctuated data. Our quantitative and qualitative analyses reveal that both approaches significantly improve punctuation robustness compared to the baseline model. Through qualitative analysis, we identified specific failure modes where NMT models fail to capture the intended meaning in the absence of punctuation. We also evaluated LLMs via zero-shot and few-shot prompting, finding that few-shot prompting improves performance. However, these models lag behind task-specific approaches in preserving meaning for punctuation-ambiguous text, highlighting the need for further research in this area.

We plan to extend this work to other Indic languages to assess whether similar qualitative patterns emerge across language families. Future work should focus on better assessment metrics that check meaning preservation and nuances similar to human judgment, and on exploring hybrid model architectures capable of handling punctuation ambiguity natively, without relying on multi-stage pipelines like multi-task learning approaches. This work opens various research directions for punctuation-robust machine translation.

\section*{Limitations}
While our study provides valuable insights into punctuation robustness, several inherent limitations bound its scope. The Viram benchmark consists of only 54 manually curated instances; although this size is sufficient for diagnostic evaluation of specific semantic ambiguities, it is not intended as a large-scale test set, with the focus deliberately placed on quality and linguistic complexity rather than volume. Our analysis is restricted to the English–Marathi language pair, and while Marathi represents a morphologically rich, low-resource Indic language, the punctuation-induced errors we observed may differ in nature and frequency for other language families or syntactic structures. 
Finally, as noted in our qualitative analysis, standard automated metrics such as BLEU and chrF are often insensitive to the subtle semantic shifts introduced by punctuation. While we supplemented these with manual inspection, the scalability of such qualitative evaluation is inherently limited due to the need for expert linguistic annotators.

\section*{Ethical Considerations}
In alignment with the ACL Ethics Policy, we provide the following disclosures regarding our data, annotation process, and potential societal impact. The English source sentences for the Viram benchmark were manually curated from a well-established linguistic resource \cite{kirkman2006punctuation}, and the fine-tuning of models in Approach 2 utilized an internal in-house corpus created by professional translators, which we plan to release publicly upon project completion to support further research. Translations for the benchmark were performed by two native Marathi speakers with advanced academic backgrounds (Master’s and PhD) in Computer Science. Annotators were fairly compensated for their specialized expertise, and all translations were developed through collaborative discussions to ensure semantic accuracy and cultural relevance. We recognize that machine translation is increasingly used in India for critical applications such as digital governance and agricultural assistance, where punctuation errors can lead to significant semantic shifts and the potential dissemination of incorrect information. While our work aims to improve model robustness and mitigate such risks, we caution that no MT system is entirely error-free, and users in sensitive domains should verify automated translations with human experts. To ensure transparency and reproducibility, we have detailed our experimental setups and prompting strategies in the Appendix and are committed to releasing the Viram diagnostic benchmark publicly to encourage more robust evaluations in Indic language technologies.

\section*{Acknowledgments}
We thank the anonymous reviewers and the meta-reviewer of LoResMT 2026 for their constructive comments and suggestions, which substantially improved this work. The first author gratefully acknowledges Shalaka Thorat for insightful discussions and significant assistance in developing the evaluation pipelines. We also thank seniors at CFILT, Satyam Kumar and Dhara Gorasiya, for their helpful feedback on the manuscript. We are grateful to the senior linguists at the CFILT Lab, Dr. Nilesh Joshi and Dr. Irawati Kulkarni, for their valuable remarks. Finally, the first author thanks his friends for their continued support and motivation.


\bibliography{custom}

\appendix

\section{More details about the \textit{Viram} Benchmark}

\subsection{Annotation Procedure}
\label{app:sec_viram_more_details}

For the creation of the \textit{Viram} benchmark translations, we hired two annotators, one pursuing a Master's degree and the other a PhD, both in the Computer Science and Engineering department. Both annotators are native speakers of Marathi. All sentences were discussed and translated collaboratively to ensure high-quality and consistent translations. The annotators received appropriate honorarium for their work.

\subsection{Data Statistics}

The human-validated English--Marathi test set contains a total of 54 instances with various punctuation marks. Commas are the most frequent, appearing 38 times, followed by colons and hyphens, each occurring 3 times. Parentheses and quotation marks appear twice each, while em dashes, question marks, semi-colons, and slashes are less frequent, with one or two occurrences. This distribution reflects the diversity of punctuation in the dataset, which may affect the complexity of translation and evaluation.

\section{Dataset construction for training models}

\subsection{Data Handling for Approach 1}
\label{app:approach_1_data}

\begin{figure*}[h]
  \centering
  \includegraphics[width=0.99\linewidth]{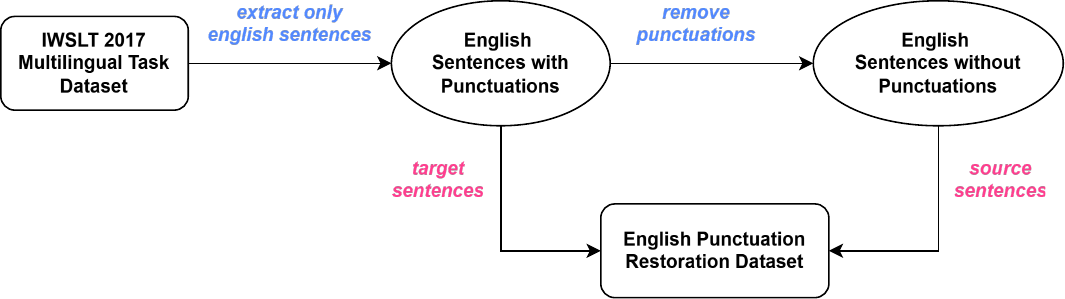}
  \caption{Data handling for the punctuation restoration task: Approach 1.}
  \label{fig:data_1}
\end{figure*}

For training the punctuation restoration models, we used the English data from the IWSLT 2017 MT challenge\footnote{\url{https://huggingface.co/datasets/IWSLT/iwslt2017}}. We considered only the English portion of the dataset, where the source sentences were stripped of punctuation and the target sentences retained the original punctuation. This setup enables the models to learn to predict and restore punctuation in English sentences. Figure~\ref{fig:data_1} illustrates the data handling process for Approach 1.

\subsection{Data Handling for Approach 2}
\label{app:approach_2_data}

For direct fine-tuning of machine translation models, we created four variants of the dataset to evaluate the effect of punctuation on translation quality: 

\begin{itemize}
    \item \textbf{Original data}: Used as a baseline without expecting any punctuation robustness. 
    \item \textbf{Data without punctuation}: All punctuation marks were removed from the source sentences to give models the ability to predict punctuation. 
    \item \textbf{Data with and without punctuation (alternate)}: Punctuation is alternately removed and retained, keeping the dataset size equal to the original. 
    \item \textbf{Data with and without punctuation (doubled)}: Each sentence is included twice, once with punctuation and once without, effectively doubling the dataset size.
\end{itemize}

\begin{figure*}[h]
  \centering
  \includegraphics[width=0.99\linewidth]{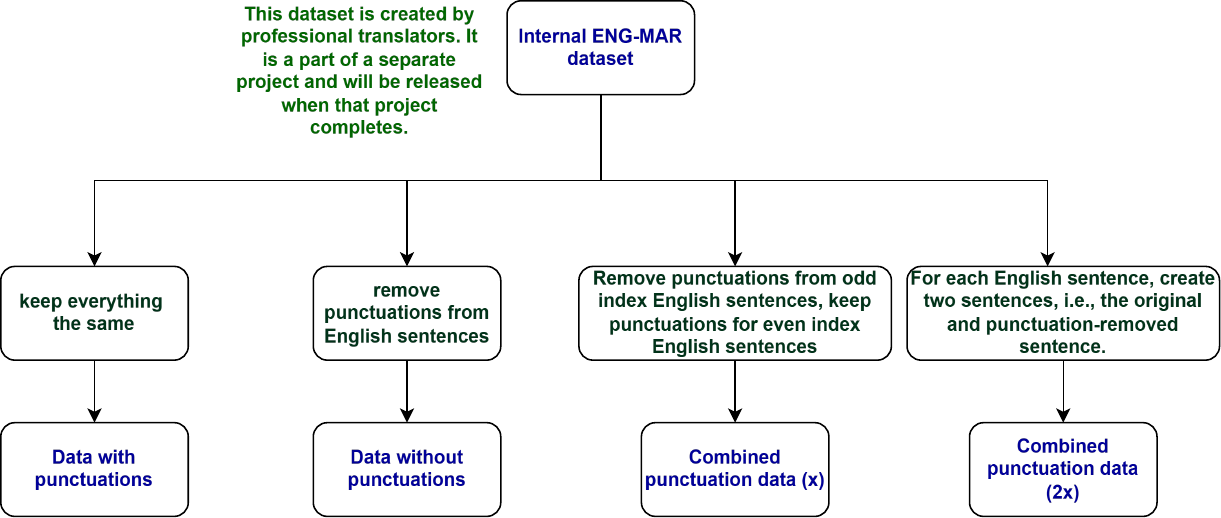}
  \caption{Data handling for the direct fine-tuning task: Approach 2.}
  \label{fig:fine_tune_mt}
\end{figure*}

Figure~\ref{fig:fine_tune_mt} shows the data handling process for Approach 2.  

\section{Statistics of the Datasets Used}
\label{app:statistics_train}

Table~\ref{tab:eng-punct} summarizes the statistics of the \texttt{english\_punctuation\_restoration} dataset. The training split contains 206,112 instances, while the validation and test splits include 888 and 8,079 instances, respectively.  

Table~\ref{tab:mar-punct} shows the dataset statistics for the internal \texttt{eng\_mar\_finetuning\_data}. The training set consists of 189,740 instances, and both the validation and test sets contain 23,717 instances each. These datasets provide the necessary coverage for training and evaluating models for punctuation restoration and English-to-Marathi fine-tuning tasks.

\begin{table}[h] \centering \begin{tabular}{l r} \hline \textbf{Split} & \textbf{Number of Instances} \\ \hline Train & 206{,}112 \\ Validation & 888 \\ Test & 8{,}079 \\ \hline \end{tabular} \caption{Dataset statistics for \texttt{english\_punctuation\_restoration}.} \label{tab:eng-punct} \end{table} 

\begin{table}[h] \centering \begin{tabular}{l r} \hline \textbf{Split} & \textbf{Number of Instances} \\ \hline Train & 189740 \\ Validation & 23717 \\ Test & 23717 \\ \hline \end{tabular} \caption{Dataset statistics for internal eng\_mar\_finetuning\_data} \label{tab:mar-punct} \end{table}





\begin{table*}[h]
\centering
\caption{Implementation details and repositories for the evaluation metrics and models.}
\label{tab:metrics_details}
\small
\resizebox{0.95\textwidth}{!}{
\begin{tabular}{@{}>{\raggedright\arraybackslash}p{0.2\linewidth}>{\raggedright\arraybackslash}p{0.4\linewidth}>{\raggedright\arraybackslash}p{0.4\linewidth}@{}}
\toprule
\textbf{Metric} & \textbf{Library / Implementation} & \textbf{Link / Repository} \\ \midrule
BLEU & Hugging Face \texttt{evaluate} (SacreBLEU) & \url{https://huggingface.co/spaces/evaluate-metric/sacrebleu} \\
chrF / chrF++ & Hugging Face \texttt{evaluate} (SacreBLEU) & \url{https://huggingface.co/spaces/evaluate-metric/chrf} \\
COMET & Hugging Face \texttt{evaluate} (Unbabel/COMET) & \url{https://github.com/Unbabel/COMET} \\
BLEURT-20 & Google Research BLEURT & \url{https://github.com/google-research/bleurt} \\
BERTScore & Hugging Face \texttt{evaluate} (MuRIL) & \url{https://huggingface.co/spaces/evaluate-metric/bertscore} \\
LaBSE & \texttt{sentence-transformers} & \url{https://huggingface.co/sentence-transformers/LaBSE} \\
MuRIL & \texttt{google/muril-base-cased} & \url{https://huggingface.co/google/muril-base-cased} \\ \bottomrule
\end{tabular}
}
\end{table*}

\section{Evaluation Metrics Used}
\label{app:metrics}

Recent advancements in Natural Language Processing (NLP), particularly in Machine Translation (MT) and cross-lingual transfer, have been driven by robust evaluation metrics and high-quality multilingual representations. This section briefly describes the evaluation metrics used in our study.

\begin{enumerate}
    \item \textbf{BLEU (Bilingual Evaluation Understudy)} \cite{papineni-etal-2002-bleu} remains one of the most widely used automatic evaluation metrics for machine translation. It computes the geometric mean of modified $n$-gram precision between a candidate translation and one or more reference translations, combined with a brevity penalty to discourage overly short outputs.

    \item \textbf{chrF++ and chrF2++} \cite{popovic-2017-chrf} are character $n$-gram–based $F$-score metrics that improve upon BLEU by capturing subword-level similarities, making them particularly effective for morphologically rich languages. While \textit{chrF++} incorporates both character and word $n$-grams, \textit{chrF2++} sets the $\beta$ parameter to 2 (i.e., an $F_2$-score), placing greater emphasis on recall than precision.

    \item \textbf{BLEURT-20} \cite{sellam-etal-2020-bleurt, sellam-etal-2020-learning} represents a shift toward learned, neural evaluation metrics. Built on a BERT-based architecture, BLEURT is pre-trained on millions of synthetic examples and fine-tuned using human judgment data. The ``-20'' checkpoint corresponds to the refined version released for the WMT 2020 Metrics shared task and exhibits strong correlation with human evaluation scores.

    \item \textbf{COMET (Cross-lingual Optimized Metric for Evaluation of Translation)} \cite{rei-etal-2020-comet} is a neural evaluation framework that leverages multilingual encoders such as XLM-RoBERTa. Unlike surface-level metrics such as BLEU, COMET jointly models the source sentence, hypothesis, and reference translation to directly predict translation quality.

    \item \textbf{LaBSE (Language-agnostic BERT Sentence Embedding)} \cite{feng-etal-2022-labse} is a dual-encoder model designed to produce language-agnostic sentence representations across 109 languages. It is trained using masked language modeling and translation ranking objectives, making it particularly effective for bitext mining and cross-lingual similarity tasks.

    \item \textbf{MuRIL (Multilingual Representations for Indian Languages)} \cite{khanuja-etal-2021-muril} is a BERT-based model tailored for the Indian linguistic landscape. Trained on 17 Indian languages and English, it incorporates both monolingual and translated/transliterated data, significantly outperforming general-purpose multilingual models (e.g., mBERT) on South Asian language tasks.
\end{enumerate}

Implementation details and repositories for the evaluation metrics and models is provided in Table \ref{tab:metrics_details}. The evaluation code used in this work follows the Indic MT Eval framework of \citet{dixit2023indicmt}.

\begin{figure*}[!ht]
\centering
\begin{tcolorbox}[width=0.75\textwidth, colframe=red!40!white, colback=red!10!white, coltitle=black!70!white, title=Prompt for Original Translation, fonttitle=\bfseries]
{
\textbf{Prompt:}\\
You are an expert linguist and translator specializing in English-to-Marathi machine translation. Translate the given English sentence into Marathi.\\

Input English: {sentence}\\

Make sure that the translation is in Devanagari Script.\\
Please provide the response in the following format:\\
Marathi Translation (Devanagari Script):
}
\end{tcolorbox}
\caption{Prompt used for original translation.}
\label{fig:prompt_original}
\end{figure*}

\begin{figure*}[!ht]
\centering
\begin{tcolorbox}[width=0.75\textwidth, colframe=green!40!white, colback=green!10!white, coltitle=black!70!white, title=Prompt for Zero-Shot Reasoning with Approach 1 (Restore then Translate), fonttitle=\bfseries]
{
\textbf{Prompt:}\\
You are an expert linguist and translator specializing in English-to-Marathi translation. You specialize in "Punctuation Restoration," resolving ambiguities caused by missing punctuation in English.\\

Steps for Analysis:\\
1. Analyze the English sentence for ambiguity.\\
2. Identify missing punctuation.\\
3. Generate the punctuated "English (Meant)" sentence.\\
4. Translate it into Marathi, ensuring the meaning is preserved.\\

Input English: {sentence}\\

Please provide the response in the following format:\\
Step 1 (Restoration): [The English (Meant) sentence]\\
Step 2 (Translation): [The Marathi translation (Devanagari Script)]\\
Reasoning: [Briefly explain your punctuation choices]
}
\end{tcolorbox}
\caption{Prompt used for zero-shot reasoning with Approach 1 (Restore then Translate)}
\label{fig:prompt_zero_cot}
\end{figure*}

\begin{figure*}[!ht]
\centering
\begin{tcolorbox}[width=0.75\textwidth, colframe=blue!40!white, colback=blue!10!white, coltitle=black!70!white, title=Prompt for Zero-Shot Translation with Approach 2 (direct translation), fonttitle=\bfseries]
{
\textbf{Prompt:}\\
You are an expert linguist and translator specializing in English-to-Marathi machine translation. Translate the given English sentence into Marathi, identifying the most logical intended meaning behind missing punctuation.\\

Input English: {sentence}\\

Make sure that the translation is in Devanagari Script.\\
Please provide the response in the following format:\\
Marathi Translation (Devanagari Script):
}
\end{tcolorbox}
\caption{Prompt used for zero-shot translation withwith Approach 2 (direct translation).}
\label{fig:prompt_zero_zero}
\end{figure*}

\begin{figure*}[!ht]
\centering
\begin{tcolorbox}[width=0.85\textwidth, colframe=orange!40!white, colback=orange!10!white, coltitle=black!70!white, title=Prompt for Few-Shot inference with Approach 1 (Restore then Translate), fonttitle=\bfseries]
{
\textbf{Prompt:}\\
You are an expert linguist and translator specializing in English-to-Marathi translation. Use punctuation restoration to resolve ambiguity.\\

Definitions:\\
1. English (Written): Unpunctuated input.\\
2. English (Meant): Restored punctuation version.\\
3. Marathi (Translation): Translation matching "English (Meant)".\\

Steps:\\
1. Analyze ambiguity.\\
2. Restore punctuation.\\
3. Translate to Marathi.\\

Some examples are as follows:\\
\begin{enumerate}
    \item \textbf{Input English:}  
    These are the components required motor brushes, bearings, and wiring.  

    \textbf{English Meant:}  
    These are the components required: motor brushes, bearings, and wiring.  

    \textbf{Marathi Translation:}  
    {\devanagarifont आवश्यक असलेले घटक खालीलप्रमाणे आहेत: मोटार ब्रशेस, बेअरिंग्ज आणि वायरिंग.}

    \item \textbf{Input English:}  
    As the machine develops the forms we use to record data from past projects will be amended.  

    \textbf{English Meant:}  
    As the machine develops, the forms we use to record data from past projects will be amended.  

    \textbf{Marathi Translation:}  
    {\devanagarifont जसजशी यंत्रणा विकसित होईल, तसतसे मागील प्रकल्पांतील डेटा रेकॉर्ड करण्यासाठी आम्ही वापरत असलेले फॉर्म्स सुधारित केले जातील.}

    \item \textbf{Input English:}  
    What we see, we believe what we hear, we register  

    \textbf{English Meant:}  
    What we see, we believe; what we hear, we register.  

    \textbf{Marathi Translation:}  
    {\devanagarifont जे पाहतो, त्यावर विश्वास ठेवतो; जे ऐकतो, त्याची नोंद घेतो.}
\end{enumerate}

Input English: {sentence}\\

Please provide the response in the following format:\\
Step 1 (Restoration): [The English (Meant) sentence]\\
Step 2 (Translation): [The Marathi translation (Devanagari Script)]\\
Reasoning: [Briefly explain your punctuation choice]
}
\end{tcolorbox}
\caption{Prompt used for few-shot inference with Approach 1 (Restore then Translate)}
\label{fig:prompt_few_cot}
\end{figure*}

\begin{figure*}[!ht]
\centering
\begin{tcolorbox}[width=0.85\textwidth, colframe=purple!40!white, colback=purple!10!white, coltitle=black!70!white, title=Prompt for Few-Shot Translation, fonttitle=\bfseries]
{
\textbf{Prompt:}\\
You are an expert linguist and translator specializing in English-to-Marathi machine translation. Translate the English sentence into Marathi, resolving ambiguity caused by missing punctuation.\\

Some examples are as follows:\\
\begin{enumerate}
    \item \textbf{Input English:}  
    These are the components required motor brushes, bearings, and wiring.  

    \textbf{English Meant:}  
    These are the components required: motor brushes, bearings, and wiring.  

    \textbf{Marathi Translation:}  
    {\devanagarifont आवश्यक असलेले घटक खालीलप्रमाणे आहेत: मोटार ब्रशेस, बेअरिंग्ज आणि वायरिंग.}

    \item \textbf{Input English:}  
    As the machine develops the forms we use to record data from past projects will be amended.  

    \textbf{English Meant:}  
    As the machine develops, the forms we use to record data from past projects will be amended.  

    \textbf{Marathi Translation:}  
    {\devanagarifont जसजशी यंत्रणा विकसित होईल, तसतसे मागील प्रकल्पांतील डेटा रेकॉर्ड करण्यासाठी आम्ही वापरत असलेले फॉर्म्स सुधारित केले जातील.}

    \item \textbf{Input English:}  
    What we see, we believe what we hear, we register  

    \textbf{English Meant:}  
    What we see, we believe; what we hear, we register.  

    \textbf{Marathi Translation:}  
    {\devanagarifont जे पाहतो, त्यावर विश्वास ठेवतो; जे ऐकतो, त्याची नोंद घेतो.}
\end{enumerate}

Input English: {sentence}\\

Make sure that the translation is in Devanagari Script.\\
Please provide the response in the following format:\\
Marathi Translation (Devanagari Script):
}
\end{tcolorbox}
\caption{Prompt used for few-shot translation with Approach 2 (direct translation)}
\label{fig:prompt_few_zero}
\end{figure*}

\section{Prompting Details}

\subsection{Original Prompt: Direct Translation without Examples}
\label{app:prompt_original}

The original prompt style instructs the model to directly translate an English sentence into Marathi without any example demonstrations or intermediate punctuation restoration. This approach tests the model's ability to perform translation with minimal guidance (see Figure~\ref{fig:prompt_original}). We used this prompt to directly input the correctly punctuated sentences to the model.

\subsection{Zero-shot Prompt: Restore Punctuation then Translate}
\label{app:prompt_zero_chain}

The zero-shot prompting strategy instructs the model to first restore punctuation in the input sentence and subsequently translate the punctuated sentence from English to Marathi. The prompt explicitly guides the model to perform punctuation restoration as an intermediate step before translation (see Figure~\ref{fig:prompt_zero_cot}).

\subsection{Zero-shot Prompt: Direct Translation} \label{app:prompt_zero_zero}
The zero-shot direct translation prompt directly instructs the model to translate punctuation-ambiguous English sentences into Marathi without any intermediate punctuation restoration step (see Figure~\ref{fig:prompt_zero_zero}).

\subsection{Three-shot Prompt: Restore Punctuation then Translate} \label{app:prompt_chain_chain}
The three-shot prompting strategy incorporates example demonstrations. Each prompt includes three input–output examples illustrating punctuation restoration followed by translation, after which the model applies the same process to a new punctuation-ambiguous sentence (see Figure~\ref{fig:prompt_few_cot}).

\subsection{Three-shot Prompt: Direct Translation}
\label{app:prompt_chain_zero}

The three-shot direct translation prompting strategy provides three input–output examples illustrating direct translation of punctuation-ambiguous English sentences into Marathi, without any intermediate punctuation restoration. The model is then asked to translate a new sentence using the same approach (see Figure~\ref{fig:prompt_few_zero}).

\section{Model Fine-tuning and Hyperparameter Tuning Details}
\label{app:fine-tune-hyper}

For machine translation experiments, we fine-tuned all models on a server equipped with four NVIDIA A100 GPUs. We conducted a comprehensive hyperparameter search, experimenting with learning rates in 
$[1\text{e-}3, 3\text{e-}3, 5\text{e-}3, 1\text{e-}4, 3\text{e-}4, 5\text{e-}4, 1\text{e-}5, 3\text{e-}5, 5\text{e-}5]$, varying the number of training epochs 
$[2, 5, 8, 10]$, and testing different batch sizes $[8, 16, 32]$. This systematic exploration allowed us to identify the most effective hyperparameter configurations for each model. The final models were selected based on their performance on the validation sets.

\end{document}